\documentclass[10pt,twocolumn,letterpaper]{article}

\usepackage{cvpr}      

\usepackage{algorithmic}
\usepackage{algorithm}
\usepackage[table]{xcolor}
\usepackage{multirow}
\usepackage{booktabs}
\usepackage{graphicx}
\usepackage{adjustbox}
\usepackage{comment}
\usepackage{amsmath}

\definecolor{cvprblue}{rgb}{0.21,0.49,0.74}
\usepackage[pagebackref,breaklinks,colorlinks,allcolors=cvprblue]{hyperref}

\def\paperID{5908} 
\def\confName{CVPR}
\def\confYear{2026}

\title{GRRE: Leveraging G-Channel Removed Reconstruction Error for Robust Detection of AI-Generated Images}

\author{Shuman He$^{1}$, Xiehua Li$^{1}$, Xioaju Yang$^{1}$, Yang Xiong$^{1}$, Keqin Li$^{1,2}$ \\
$^{1}$ College of Cyber Science and Technology, Hunan University\\
$^{2}$ Department of Computer Science, State University of New York at New Paltz\\
{\tt\small annahe@hnu.edu.cn}
\and
\\
}

\begin{document}
\maketitle
\begin{abstract}
The rapid progress of generative models, particularly diffusion models and GANs, has greatly increased the difficulty of distinguishing synthetic images from real ones. Although numerous detection methods have been proposed, their accuracy often degrades when applied to images generated by novel or unseen generative models, highlighting the challenge of achieving strong generalization. To address this challenge, we introduce a novel detection paradigm based on channel removal reconstruction. Specifically, we observe that when the green (G) channel is removed from real images and reconstructed, the resulting reconstruction errors differ significantly from those of AI-generated images. Building upon this insight, we propose {\bfseries\itshape G}-channel {\bfseries\itshape R}emoved {\bfseries\itshape R}econstruction {\bfseries\itshape E}rror ({\bfseries\itshape GRRE}), a simple yet effective method that exploits this discrepancy for robust AI-generated image detection. Extensive experiments demonstrate that GRRE consistently achieves high detection accuracy across multiple generative models, including those unseen during training. Compared with existing approaches, GRRE not only maintains strong robustness against various perturbations and post-processing operations but also exhibits superior cross-model generalization. These results highlight the potential of channel-removal-based reconstruction as a powerful forensic tool for safeguarding image authenticity in the era of generative AI.
\end{abstract}    
\begin{figure}[t]
    \centering
    \includegraphics[width=\linewidth]{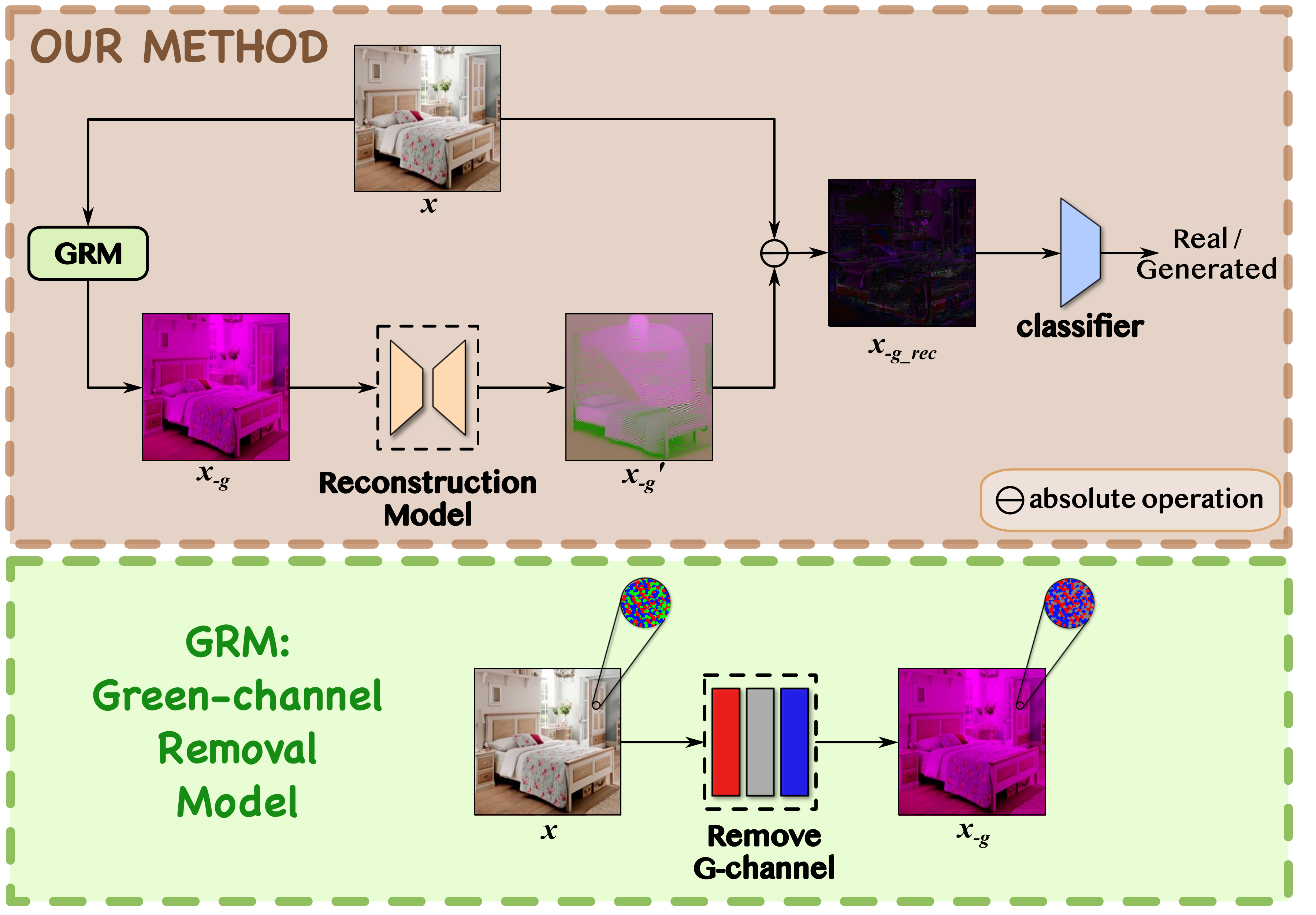}
    \caption{
        Overview of the proposed \textbf{G-channel Removed Reconstruction Error (GRRE)} framework. 
        Given an input image $x$, the Green-channel Removal Model (GRM) removes the green channel to obtain $x_{-g}$.
        The diffusion-based reconstruction model generates a restored image $x_{-g}'$, and their difference forms the $x_{-g\_rec}$ map. 
        This map captures channel-sensitive reconstruction discrepancies, which are subsequently leveraged for distinguishing real and AI-generated images.
    }
    \label{fig:grre_pipeline}
\end{figure}
\section{Introduction}
\label{sec:intro}

Recent advances in generative modeling, particularly Generative Adversarial Networks (GANs) \cite{GANs,ProGAN,StyleGAN,StyleGAN2} and diffusion models\cite{DDPMs,iDDPM,LDMs}, have enabled the creation of highly realistic and diverse images, with wide applications in content editing, digital art, and creative design \cite{Diffusion-basedVisualArtCreation,GlazeProtectingArtists,JieHuaStyleFeatureExtracting,PDANetDiffusionPainting}. While these technologies offer remarkable benefits, they also introduce critical risks, as synthetic media can be misused for identity forgery and misinformation, thereby amplifying concerns over content authenticity \cite{DHSDeepfake2023,DeepfakeUnveiled,DeepfakeIdentityTheft,DeepfakeDetection&MultimediaForensics}. Consequently, AI-generated image detection has become an urgent research focus, requiring methods that are not only accurate but also robust and generalizable to rapidly evolving generative techniques.

Despite considerable progress, existing approaches to AI-generated image detection face notable limitations. Reconstruction-based methods \cite{DIRE,LaRE2} exploit discrepancies revealed through generative reconstructions; however, they are often computationally expensive and show degraded performance when applied to unseen architectures. Representation-based methods \cite{FIRE,PatchCraft,CSFD,HFMF}, relying on frequency artifacts or texture features, achieve high in-domain performance but exhibit limited transferability when applied to images from unseen distributions. Watermarking and fingerprinting strategies \cite{GANFingerprints,TreeRingWatermarks,StableSignature} provide reliable provenance tracking when embedded during generation. However, their applicability is limited because not all generative models incorporate such mechanisms, especially in open-source environments. Moreover, large-scale benchmarks \cite{GenImage,FaceForensics++,FakeBench} reveal that the performance of most methods drops significantly when evaluated across diverse generators, highlighting the urgent need for more robust and generalized solutions.

To overcome the limitations of existing detectors, we propose G-channel Removed Reconstruction Error (GRRE), a framework that exploits the critical role of the green channel in image quality and feature representation to effectively distinguish real images from synthetic ones.
Prior studies have demonstrated that the green channel contributes most to luminance, texture, and structural detail, and its removal causes significant degradation in image fidelity \cite{ColoeImageDenoisingGreenChannelPrior,ImageDenoisingGreenChannel,JointDenoisingDemosaickingGreenChannel,FoundationVision}. Building on this observation, GRRE removes the green channel from input images, reconstructs the corrupted inputs via a diffusion model, and measures the resulting reconstruction error. Since real images typically exhibit larger discrepancies while synthetic ones align more closely with the model’s manifold, our approach leverages the intrinsic properties of color channels to distinguish real from synthetic images, yielding enhanced robustness and strong generalization across generative models. The overall framework of GRRE is illustrated in Figure\ref{fig:grre_pipeline}.

In summary, the contributions of our work are threefold:

\begin{itemize}[leftmargin=*, align=parleft]
  \item We are the first to introduce G-channel Removed Reconstruction Error (GRRE), a novel detection framework that removes the green-channel and employs diffusion-based reconstruction to amplify the intrinsic discrepancies between real and synthesized images.
  \item We propose a reconstruction-based representation that leverages the perceptual dominance of the green channel in visual sensitivity and image fidelity. This design enables consistent detection of subtle reconstruction inconsistencies across diverse generative models, achieving robust and generalizable detection performance.
  \item Moreover, we conduct extensive experiments across multiple generative models and datasets,
        demonstrating that our method achieves superior robustness and strong cross-model generalization capability.
\end{itemize}


\section{Related Work}
\label{sec:formatting}

\subsection{Image Generation}

The field of image generation has advanced rapidly, evolving from adversarial to diffusion-based paradigms.
Generative Adversarial Networks (GANs)~\cite{GANs} pioneered adversarial training, followed by DCGAN~\cite{DCGAN}, ProGAN~\cite{ProGAN}, and StyleGAN~\cite{StyleGAN, StyleGAN2}, which enabled high-resolution, photorealistic synthesis with controllable styles.
Recently, diffusion models have surpassed GANs in fidelity and diversity.
Denoising Diffusion Probabilistic Models (DDPMs)~\cite{DDPMs} and their variants~\cite{iDDPM, SDEs, DMsGAN} establish a new generation paradigm, while Latent Diffusion Models (LDMs)~\cite{LDMs} improve efficiency through latent-space operations.
Large-scale text-to-image systems such as DALL·E~\cite{DALL-E, dalle2, dalle3}, Imagen~\cite{Imagen}, and Stable Diffusion~\cite{LDMs, SDXL} have achieved unprecedented realism.
Recent models such as PixArt-$\Sigma$~\cite{PixArt}, SD3~\cite{SD3, SD3-Turbo}, Sana~\cite{Sana}, and FlowTok~\cite{FlowTok} further improve generation quality and efficiency while raising new challenges for detecting AI-generated images.

\subsection{AI-Generated Image Detection}
Existing approaches for AI-generated image detection can be broadly grouped into three paradigms: reconstruction-based methods, representation-based methods, and watermarking/fingerprinting strategies. In addition, large-scale datasets and benchmarks have been developed to systematically evaluate detection performance across diverse generative models.

Reconstruction-based approaches exploit discrepancies between real and synthetic images revealed through generative model reconstructions. Representative works include DIRE~\cite{DIRE}, which measures diffusion reconstruction error, and LaRE2~\cite{LaRE2}, which leverages latent reconstruction error. These methods demonstrate the potential of reconstruction signals for detection, though robust cross-model generalization remains challenging.

Representation-based approaches directly learn discriminative features such as frequency cues or local textures from images without explicit reconstruction. Notable examples include FIRE~\cite{FIRE}, which integrates frequency decomposition into a reconstruction method; PatchCraft~\cite{PatchCraft}, which emphasizes texture patch selection; and CSFD~\cite{CSFD}, which further exploits spectral information. Methods such as HFMF~\cite{HFMF} adopt hierarchical multi-stream fusion for enhanced robustness. Earlier works also reveal GAN fingerprints  \cite{GANFingerprints} and frequency artifacts \cite{FrequencyFeaturesGanDetection}, underscoring the importance of spectral and structural cues in representation learning.

Another important direction is watermarking and fingerprinting, which focus on content attribution and traceability. GAN Fingerprints \cite{GANFingerprints} showed that different architectures leave distinct model-specific artifacts. More recent approaches, such as Tree-Ring Watermarks \cite{TreeRingWatermarks} and Stable Signature~\cite{StableSignature}, embed imperceptible identifiers into generated outputs to enable provenance verification. While effective in cooperative settings, these methods often struggle in open-source or unmarked generative environments.

Although existing detection methods have achieved notable progress, each category faces inherent limitations. \textbf{Reconstruction-based approaches} often suffer from high computational cost and limited cross-model generalization. \textbf{Representation-based methods} may overfit to specific datasets and fail under distribution shifts. \textbf{Watermarking and fingerprinting techniques} rely on model cooperation, making them less effective in open-source or unmarked scenarios.

To overcome these challenges, we propose GRRE (G-channel Removed Reconstruction Error), which leverages the critical role of the green channel in image quality and feature representation. By removing the green channel and reconstructing the modified image to obtain reconstruction errors, GRRE captures the significant discrepancy between real and generated images, achieving high accuracy, strong robustness, and superior generalization in distinguishing authentic from synthetic content.


\section{Proposed Method}

In this section, we present the proposed detection framework based on the G-channel Removed Reconstruction Error (GRRE).
The key idea of our method is to exploit the reconstruction discrepancy induced by removing the green channel from an image and analyzing its reconstruction behavior through diffusion models.
To this end, we organize the methodology into three parts.
First, we review the RGB image representation and introduce the channel removal operation, which forms the basis of our approach.
Second, we describe the diffusion-based reconstruction process and formally define the GRRE as a discriminative feature to distinguish between real and AI-generated images.
Finally, we present the design of the classification network and the loss function used, which together enable the effective exploitation of GRRE features for robust binary classification.
Together, these components constitute a comprehensive methodological foundation for reliable AI-generated image detection.

\subsection{RGB Image Model}
RGB (Red, Green, Blue) is one of the most widely used color models, commonly applied in digital imaging, displays, cameras, and computer vision tasks. As an additive color model, RGB represents nearly the entire gamut of colors reproducible in images by combining the three primary color channels (R, G, and B) at varying intensities. In digital systems, an RGB image is typically stored as a three-channel matrix, where each channel corresponds to a grayscale image encoding the intensity of a single color component. Formally, a digital image $I(x,y)$ of size $H \times W$ can be expressed as a three-dimensional tensor:
\begin{equation}
    I(x,y) = [R(x,y), G(x,y), B(x,y)]
\end{equation}
where $R(x,y)$ denotes the red component at pixel $(x,y)$, $G(x,y)$ the green component, and $B(x,y)$ the blue component. The displayed color at each pixel $C(x,y)$  is determined by a linear combination of the three channels:
\begin{equation}
    C(x,y) = \alpha_R \cdot R(x,y) + \alpha_G \cdot G(x,y) + \alpha_B \cdot B(x,y)
\end{equation}
where $\alpha_R, \alpha_G, \alpha_B \in [0,1] $ denote the relative intensity coefficients of each channel. 

For a given RGB image $I(x,y)$, its G-channel Removal version is defined as:
\begin{equation}
    I_{-G}(x,y) = [R(x,y), 0, B(x,y)]
\end{equation}
The Green-channel Removal(GRM) operation discards the green component. The corresponding implementation can be expressed as follows:
\begin{algorithm}[H]
\caption{Green-channel Removal for RGB Images}\label{alg:GRemoval}
\begin{algorithmic}[1]
\REQUIRE RGB image $I \in \mathbb{R}^{H \times W \times 3}$
\ENSURE The G-channel Removed image $I_{-G}$

\FOR{$x = 1$ to $H$}
    \FOR{$y = 1$ to $W$}
        \STATE $R \leftarrow I(x,y,1)$ \hfill \COMMENT{Red channel}
        \STATE $B \leftarrow I(x,y,3)$ \hfill \COMMENT{Blue channel}
        \STATE $I_{-G}(x,y) \leftarrow [R, 0, B]$ \hfill \COMMENT{Discard green channel}
    \ENDFOR
\ENDFOR

\RETURN $I_{-G}$
\end{algorithmic}
\end{algorithm}

\subsection{GRRE}
The core idea of our method is to exploit the reconstruction discrepancy induced by removing the green channel from an input image. This process consists of two main steps: {\bf(i) image reconstruction} based on diffusion probabilistic models (DDPM), and {\bf(ii) computation of the reconstruction error} between the original and reconstructed images.
\subsubsection{Image Reconstruction with Diffusion Models}
Diffusion probabilistic models (DDPMs)~\cite{DDPMs} generate images by progressively denoising a latent variable to recover an image from noise. Given an original image $x_0$, the forward diffusion process adds Gaussian noise to it step by step:
\begin{equation}
    q(x_t \mid x_{t-1}) = \mathcal{N}(x_t; \sqrt{1-\beta_t} \, x_{t-1}, \beta_t I),
\end{equation}
where $x_t$ represents the noisy image at timestep $t$.
$\mathcal{N}(\cdot)$ denotes a multivariate Gaussian distribution.
$\beta_t \in (0, 1)$ is a variance schedule parameter that controls the amount of noise added at each step.
$I$ is the identity matrix, ensuring isotropic covariance.
$t \in \{1, 2, \ldots, T\}$ indexes the diffusion step.
The coefficient $\sqrt{1 - \beta_t}$ retains the signal proportion from the previous step $x_{t-1}$.
This formulation defines the forward noising process in DDPMs, which progressively transforms the data distribution into a standard normal distribution for subsequent denoising-based generation.
$q(x_t \mid x_{t-1})$ models the \textit{forward transition probability} from $x_{t-1}$ to $x_t$.
The closed-form expression of the noised image at step $t$ can be written as:
\begin{equation}
    q(x_t \mid x_0) = \mathcal{N}\left(x_t; \sqrt{\bar{\alpha}_t} \, x_0, (1-\bar{\alpha}_t) I \right),
\end{equation}
with $\bar{\alpha}_{t} \;=\; \prod_{s=1}^{t} (1-\beta_{s})$. The reverse process aims to reconstruct the original image by progressively denoising from a noisy latent variable to $x'_0$:
\begin{equation}
p_\theta(x_{t-1} \mid x_t) = \mathcal{N}\left(x_{t-1}; \mu_\theta(x_t, t), \Sigma_\theta(x_t, t)\right),
\end{equation}
where $\mu_\theta$ and $\Sigma_\theta$ are learned denoising parameters parameterized by a neural network with parameters $\theta$.

In our setting, the green channel is first removed from the RGB image to obtain $I_{-G}$. This corrupted image is then used as input for the diffusion-based reconstruction process. The reconstructed image after iterative denoising is given by:
\begin{equation}
    \hat{I}_{-G} = \mathcal{R}_{\theta}(I_{-G}),
\end{equation}
where $\mathcal{R}_{\theta}(\cdot)$ denotes the reconstruction operator parameterized by a denoising diffusion probabilistic model (DDPM) with learnable parameters $\theta$. Specifically, $\mathcal{R}_{\theta}$ performs the reverse diffusion process, progressively removing noise from the corrupted input $I_{-G}$ through a sequence of denoising steps. At each timestep $t$, the model predicts the clean image estimate by approximating the conditional posterior $p_{\theta}(x_{t-1}\mid x_t)$. After iterating from timestep $T$ down to $0$, the operator outputs the reconstructed image $\hat{I}_{-G}$.
\subsubsection{Reconstruction Error Computation}
After obtaining the reconstructed image $\hat{I}_{-G}$, we compute the reconstruction error by directly measuring the difference between the original image $I$ and $\hat{I}_{-G}$. To formally define this process, we introduce the reconstruction error as follows:
\begin{equation}
    \mathrm{GRRE}(x_0) = \big\lvert I(x_0) - \hat{I}_{-G}(x_0) \big\rvert,
\end{equation}
where $I(x_0)$ denotes the original input image, $\hat{I}_{-G}(x_0)$ represents the reconstructed image obtained after removing the green channel and applying the diffusion-based reconstruction operator, and $\mathrm{GRRE}(x_0)$ quantifies the discrepancy between them.

The computed $\mathrm{GRRE}(x_0)$ highlights the distributional differences between real and AI-generated images: real images typically produce larger reconstruction discrepancies due to the loss of critical structural information in the green channel, while generated images exhibit smoother reconstructions that mask such discrepancies. This property makes $\mathrm{GRRE}(x_0)$ an effective discriminative feature for AI-generated image detection.

To make the computation procedure more transparent and reproducible, we summarize the reconstruction error calculation as a pseudocode description, shown in Algorithm~\ref{alg:GRRE}.
\begin{algorithm}[H]
\caption{Reconstruction Error Computation}\label{alg:GRRE}
\begin{algorithmic}[1]
\REQUIRE Original image $I(x_0)$, reconstructed image $\hat{I}_{-G}(x_0)$
\ENSURE Reconstruction error $\mathrm{GRRE}(x_0)$
\STATE Compute pixel-wise difference: $D \leftarrow I(x_0) - \hat{I}_{-G}(x_0)$
\STATE Aggregate difference to obtain reconstruction error:
       $\mathrm{GRRE}(x_0) \leftarrow \big\lvert D \big\lvert$
\RETURN $\mathrm{GRRE}(x_0)$
\end{algorithmic}
\end{algorithm}

\subsection{Classification and Loss}
To discriminate between real and AI-generated images, we employ a convolutional neural network classifier trained on the GRRE maps. Specifically, we adopt a ResNet-50 backbone, which has demonstrated strong generalization ability in various image analysis tasks. The final fully connected layer is replaced with a single-node output layer for binary classification. Given an input error map $\mathrm{GRRE}(x_0)$, the classifier outputs a logit value:
\begin{equation}
    z = f(\mathrm{GRRE}(x_0)),
\end{equation}
where $f(\cdot)$ denotes the ResNet-50 feature extractor. 
We employ the Binary Cross-Entropy with Logits Loss (BCEWithLogitsLoss), which incorporates the sigmoid operation internally for improved numerical stability. 
Given a dataset with input-label pairs $\{\mathrm{GRRE}(x_0^i), y^i\}_{i=1}^N$, where $y^i \in \{0,1\}$ denotes the ground-truth label (0 for real, 1 for generated), 
the loss function is defined as:
\begin{equation}
    \mathcal{L} = \frac{1}{N} \sum_{i=1}^{N} 
    \Big[ \log\!\big(1 + e^{-z^i}\big) + (1 - y^i)z^i \Big], 
\end{equation}
\text{where } $z^i = f_\theta(\mathrm{GRRE}(x_0^i))$ represents the \textit{logit} output of the network before the sigmoid activation, corresponding to the input error map $\mathrm{GRRE}(x_0^i)$.
This formulation ensures numerical stability and encourages the network to output high logits for generated images ($y=1$) and low logits for real images ($y=0$).

During training, we apply standard data augmentation strategies, including random cropping and horizontal flipping, to improve generalization. The optimizer is AdamW with weight decay for regularization, and the learning rate is dynamically adjusted using a ReduceLROnPlateau scheduler. Furthermore, early stopping is employed to prevent overfitting, and automatic mixed precision (AMP) is used to accelerate training on GPU while reducing memory consumption.
\section{Experiments}
\subsection{Setup}
\noindent\textbf{Datasets.} All experiments are conducted on the DiffusionForensics dataset~\cite{DIRE}, a comprehensive benchmark designed for detecting AI-generated images. It consists of three sub-datasets—CelebA-HQ~\cite{ProGAN}, ImageNet~\cite{imagenet}, and LSUN-Bedroom~\cite{LSUN}—each containing real images and those synthesized by multiple diffusion or text-to-image models. Specifically, the CelebA-HQ subset includes synthetic faces generated by DALL·E 2~\cite{dalle2}, IF~\cite{if}, Midjourney, and Stable Diffusion v2(SD-v2)~\cite{SDXL}. The ImageNet subset comprises samples generated by ADM~\cite{DMsGAN} and Stable Diffusion v1(SD-v1)~\cite{SDXL}. The LSUN-Bedroom subset covers a wider range of generation paradigms, including ADM, DDPM~\cite{DDPMs}, iDDPM~\cite{iDDPM}, PNDM~\cite{pndm}, SD-v1, SD-v2, StyelGAN\cite{StyleGAN}, LDM~\cite{LDMs}, VQ-Diffusion~\cite{vqdiffusion}, IF, DALL·E 2 and Midjourney. All images are resized to a resolution of $256 \times 256$ and are split into official training, validation, and testing sets as described in~\cite{DIRE}. 

\noindent\textbf{Evaluation metrics.} Following previous generated-image detection methods~\cite{DIRE,aeroblade,fakeinversion,FIRE} , we evaluate the detection performance using three standard metrics widely adopted in image forensics and binary classification: Accuracy (Acc), Area Under the ROC Curve (AUC), and Average Precision (AP). ACC measures the overall proportion of correctly classified real and generated images. AUC reflects the model’s discrimination capability across all decision thresholds by integrating the Receiver Operating Characteristic (ROC) curve. AP computes the area under the precision–recall curve and highlights the model’s robustness to threshold selection, particularly in imbalanced scenarios. The threshold for computing accuracy is set to 0.5~\cite{DIRE}.

\noindent\textbf{Baselines.} We compare our method with several state-of-the-art reconstruction-based detectors. All baselines are reproduced using their official open-source implementations and training protocols, and are trained and tested on our datasets under identical conditions for a fair comparison.
1) \textbf{DIRE}~\cite{DIRE} introduces a diffusion-based reconstruction error forensics framework, where a pretrained diffusion model reconstructs each input and the pixel-wise reconstruction error is used to distinguish real and synthetic images. 
2) \textbf{AEROBLADE}~\cite{aeroblade} proposes a unified autoencoder-based detector that measures reconstruction distance in latent space and achieves strong generalization across unseen generative models. 
3) \textbf{FakeInversion}~\cite{fakeinversion} leverages latent inversion consistency between real and generated images by comparing reconstruction fidelity in diffusion and GAN domains, effectively capturing inversion discrepancies for fake detection. 
4) \textbf{FIRE}~\cite{FIRE} models the frequency-domain reconstruction error of diffusion models and demonstrates robust detection under diverse post-processing perturbations and cross-model settings.

\begin{table*}[t]
\caption{
\textbf{Comparison of GRRE (ours) with existing detectors.}
Each cell reports ACC/AUC (\%) scores. 
}
\centering
\renewcommand{\arraystretch}{1.1}
\setlength{\tabcolsep}{3pt}
\resizebox{0.95\textwidth}{!}{
\begin{tabular}{l|l|l ccccc ccccc}
\toprule
\multirow{2}{*}{\textbf{Eval set}} &
\multicolumn{2}{c}{\shortstack{\textbf{Train set w.}\\ \textbf{Gen. method}}} &
\multicolumn{5}{c}{\raisebox{0.8ex}[0pt][0pt]{\centering\textbf{CelebA-HQ w. SD-v2}}} &
\multicolumn{5}{c}{\raisebox{0.8ex}[0pt][0pt]{\centering\textbf{LSUN-B. w. StyleGAN}}} \\
\cmidrule(lr){2-13}
& \shortstack{\textbf{Gen.}\\ \textbf{method}} & \raisebox{0.8ex}[0pt][0pt]{\textbf{Model}} &
\raisebox{0.8ex}[0pt][0pt]{AEROBLADE\cite{aeroblade}} & \raisebox{0.8ex}[0pt][0pt]{FakeInversion\cite{fakeinversion}} & \raisebox{0.8ex}[0pt][0pt]{DIRE\cite{DIRE}} & \raisebox{0.8ex}[0pt][0pt]{FIRE\cite{FIRE}} & \raisebox{0.8ex}[0pt][0pt]{\textbf{GRRE(Ours)}} &
\raisebox{0.8ex}[0pt][0pt]{AEROBLADE} & \raisebox{0.8ex}[0pt][0pt]{FakeInversion} & \raisebox{0.8ex}[0pt][0pt]{DIRE} & \raisebox{0.8ex}[0pt][0pt]{FIRE} & \raisebox{0.8ex}[0pt][0pt]{\textbf{GRRE(Ours)}} \\
\midrule
\multirow{4}{*}{Celeba-HQ}
& \multicolumn{2}{l}{DALLE·2\cite{dalle2}} & 73.47/72.28 & 67.93/61.81 & 99.60/100.0 & 86.07/99.70 & \cellcolor{blue!10}\textbf{100.0/100.0} & 73.47/72.28 & 66.93/80.15 & 89.67/89.70 & 66.67/49.70 & \cellcolor{blue!10}\textbf{100.0/100.0} \\
& \multicolumn{2}{l}{IF\cite{if}} & 99.35/99.97 & 55.60/83.49 & 99.00/100.0 & 96.85/99.87 & \cellcolor{blue!10}\textbf{99.95/100.0} & 99.35/99.97 & 55.95/98.51 & 92.80/96.22 & 50.00/39.35 & \cellcolor{blue!10}\textbf{100.0/100.0} \\
& \multicolumn{2}{l}{Mid.} & 95.45/97.76 & 90.91/82.79 & 99.82/100.0 & 99.64/99.88 & \cellcolor{blue!10}\textbf{100.0/100.0} & 95.45/97.76 & 91.00/96.92 & 93.64/99.74 & 90.91/30.91 & \cellcolor{blue!10}\textbf{100.0/100.0} \\
& \multicolumn{2}{l}{SD-v2\cite{SDXL}} & 69.65/75.73 & 99.70/99.99 & 99.90/100.0 & 99.95/100.0 & \cellcolor{blue!10}\textbf{100.0/100.0} & 69.65/75.73 & 55.60/93.19 & 94.35/98.58 & 50.00/55.99 & \cellcolor{blue!10}\textbf{100.0/100.0} \\
\midrule
\multirow{2}{*}{ImageNet}
& \multicolumn{2}{l}{ADM\cite{DMsGAN}} & 67.71/74.21 & 54.00/60.76 & 78.02/96.06 & 53.15/54.65 & \cellcolor{blue!10}\textbf{91.60/98.61} & 67.71/74.21 & 53.14/95.98 & \cellcolor{blue!10}\textbf{93.15}/99.08 & 50.07/56.81 & 85.17/\textbf{100.0} \\
& \multicolumn{2}{l}{SD-v1\cite{SDXL}} & \cellcolor{blue!10}\cellcolor{blue!10}\textbf{94.41}/97.53 & 41.49/53.82 & 83.04/97.53 & 47.04/73.54 & 91.69/\textbf{98.59} & \cellcolor{blue!10}\textbf{94.41}/97.53 & 50.05/48.19 & 93.50/99.21 & 33.43/52.59 & 85.17/\textbf{99.76} \\
\midrule
\multirow{11}{*}{LSUN-B.}
& \multicolumn{2}{l}{ADM} & 72.45/77.16 & 58.00/66.43 & 89.40/98.98 & 56.35/60.39 & \cellcolor{blue!10}\textbf{98.65/99.93} & 72.45/77.16 & 98.25/99.82 & 95.60/99.81 & 77.35/96.25 & \cellcolor{blue!10}\textbf{100.0/100.0} \\
& \multicolumn{2}{l}{DDPM\cite{DDPMs}} & 78.85/87.06 & 58.31/57.61 & 91.69/99.09 & 57.98/47.91 & \cellcolor{blue!10}\textbf{98.19/99.86} & 78.85/87.06 & 59.56/94.25 & 99.04/99.96 & 77.32/81.07 & \cellcolor{blue!10}\textbf{100.0/100.0} \\
& \multicolumn{2}{l}{IDDPM\cite{iDDPM}} & 71.05/76.95 & 55.90/63.75 & 83.50/98.55 & 52.20/52.36 & \cellcolor{blue!10}\textbf{98.10/99.89} & 71.05/76.95 & 98.85/99.84 & 97.30/99.95 & 79.55/96.79 & \cellcolor{blue!10}\textbf{100.0/100.0} \\
& \multicolumn{2}{l}{PNDM\cite{pndm}} & 61.15/65.73 & 47.60/21.26 & 70.60/92.71 & 49.25/45.63 & \cellcolor{blue!10}\textbf{95.75/99.61} & 61.15/65.73 & 50.40/88.09 & 95.20/99.52 & 65.80/88.04 & \cellcolor{blue!10}\textbf{100.0/100.0} \\
& \multicolumn{2}{l}{SD-v1} & 99.80/99.99 & 47.10/13.49 & 95.45/99.77 & 48.80/38.16 & \cellcolor{blue!10}\textbf{99.75/100.0} & 99.80/99.99 & 49.75/38.75 & 99.95/100.0 & 55.40/96.67 & \cellcolor{blue!10}\textbf{100.0/100.0} \\
& \multicolumn{2}{l}{SD-v2} & 69.90/75.10 & 72.40/86.33 & 97.35/99.84 & 67.90/79.80 & \cellcolor{blue!10}\textbf{100.0/100.0} & 69.90/75.10 & 49.75/31.60 & 99.90/100.0 & 49.85/59.31 & \cellcolor{blue!10}\textbf{100.0/100.0} \\
& \multicolumn{2}{l}{LDM\cite{LDMs}} & 99.15/99.83 & 51.55/57.28 & 98.70/99.93 & 87.50/94.51 & \cellcolor{blue!10}\textbf{99.95/100.0} & 99.15/99.83 & 61.60/98.60 & 99.70/100.0 & 49.95/63.65 & \cellcolor{blue!10}\textbf{100.0/100.0} \\
& \multicolumn{2}{l}{VQD\cite{vqdiffusion}} & 83.70/90.30 & 49.85/58.05 & 97.45/99.84 & 59.65/63.65 & \cellcolor{blue!10}\textbf{99.95/100.0} & 83.70/90.30 & 71.70/98.94 & 99.90/100.0 & 50.00/67.24 & \cellcolor{blue!10}\textbf{100.0/100.0} \\
& \multicolumn{2}{l}{IF} & 96.35/98.98 & 58.50/81.23 & 99.30/99.98 & 75.10/82.57 & \cellcolor{blue!10}\textbf{100.0/100.0} & 96.35/98.98 & 64.65/98.63 & 99.80/100.0 & 49.85/56.54 & \cellcolor{blue!10}\textbf{100.0/100.0} \\
& \multicolumn{2}{l}{DALLE·2} & 64.27/62.86 & 63.07/35.26 & 99.53/99.97 & 69.87/50.70 & \cellcolor{blue!10}\textbf{99.93/100.0} & 64.27/62.86 & 66.33/37.40 & 99.73/99.99 & 66.53/69.25 & \cellcolor{blue!10}\textbf{100.0/100.0} \\
& \multicolumn{2}{l}{Mid.} & 66.82/77.92 & 85.64/31.88 & 98.91/98.70 & 88.73/45.63 & \cellcolor{blue!10}\textbf{100.0/100.0} & 66.82/77.92 & 90.45/24.90 & 99.91/100.0 & 90.64/75.14 & \cellcolor{blue!10}\textbf{100.0/100.0} \\
\midrule
\multicolumn{3}{l}{\textbf{Average}} & 80.21/84.08 & 62.21/59.72 & 93.02/98.87 & 70.35/69.92 & \cellcolor{blue!10}\textbf{98.44/99.79} & 80.21/84.08 & 66.70/77.87 & 96.66/98.93 & 62.89/68.14 & \cellcolor{blue!10}\textbf{98.26/99.99} \\
\bottomrule
\end{tabular}
}
\label{tab:comparison}
\end{table*}

\subsection{Comparison to Existing Detectors}
\paragraph{Experimental Design.}
We compare our proposed \textbf{GRRE} framework with four representative baselines: AEROBLADE, FakeInversion, DIRE, and FIRE.  
In our experiments, we utilize the complete DiffusionForensics dataset to evaluate both \textbf{Intra-Dataset Cross-Model} and \textbf{Cross-Dataset} generalization: 
(1)\textbf{CelebA-HQ $\rightarrow$ Others:} we train the detector on the CelebA-HQ subset with images generated by SD-v2(20k fake, 20k real) and test on all remaining subsets; 
(2)\textbf{LSUN-Bedroom $\rightarrow$ Others:} we train on the LSUN-Bedroom subset with images generated by StyleGAN(40k fake, 40k real) and test on all remaining subsets.
This protocol ensures that the training and testing sets are disjoint across both semantic domains and generative models, providing a rigorous evaluation of the model's generalization capability.
Both experiments adopt consistent preprocessing pipelines, evaluation metrics (\textbf{ACC} and \textbf{AUC}), and backbone architectures where applicable.  
This dual-setting design allows us to comprehensively examine both in-domain and cross-domain generalization capabilities of each detector.

\paragraph{Intra-Dataset Cross-Model Results.}
To evaluate the within-dataset generalization ability, we train detectors on CelebA-HQ(SD-v2) and LSUN-Bedroom(StyleGAN), and test them across all generation models within the same dataset.
The results are summarized in \cref{tab:comparison}. Our proposed GRRE achieves consistent and superior performance on all test generators, significantly outperforming AEROBLADE, FakeInversion, DIRE, and FIRE.

When trained on CelebA-HQ (SD-v2) and evaluated across DALLE·2, IF, Midjourney, and SD-v2 test sets, GRRE attains nearly perfect results, reaching 100 / 100 ACC/AUC in most cases.
Compared with DIRE and FIRE, which already achieve strong in-dataset generalization, GRRE still provides a noticeable improvement.
This clearly indicates that the G-channel removed reconstruction mechanism enhances the separability between real and generated content, even when both the training and testing datasets share similar semantic distributions.
These results verify that GRRE captures generator-agnostic reconstruction inconsistencies and achieves near-saturation accuracy within the same dataset.

A similar trend is observed on LSUN-Bedroom (StyleGAN), where the generative diversity and textural complexity are much higher.
Despite the broader variation across diffusion-based and GAN-based models (ADM, DDPM, IDDPM, PNDM, SD-v1/v2, LDM, VQD, IF, DALLE·2, Midjourney), GRRE achieves 100 / 100 ACC/AUC for all 11 generators, demonstrating a clear advantage over previous methods.
While FIRE and DIRE occasionally exceed 99 \% AUC, their performance fluctuates noticeably between diffusion and GAN families, suggesting limited adaptability.
In contrast, GRRE produces stable results across all architectures, demonstrating exceptional robustness and uniformity in intra-dataset generalization.
This consistency implies that removing the green channel before reconstruction effectively exposes latent structural artifacts that are common across different generator types, rather than relying on dataset-specific cues.

\paragraph{Cross-Dataset Generalization.}
To further evaluate the transferability of the proposed method, we conduct cross-dataset generalization experiments under two transfer settings:
(1) models trained on CelebA-HQ(SD-v2) are tested on ImageNet and LSUN-Bedroom, and
(2) models trained on LSUN-Bedroom(StyleGAN) are tested on CelebA-HQ and ImageNet.
This setting is highly challenging, as both scene semantics and generative priors vary substantially across datasets.

As shown in \cref{tab:comparison}, when trained on CelebA-HQ (SD-v2) and transferred to ImageNet and LSUN-Bedroom, GRRE consistently achieves the highest detection accuracy and AUC across all generation models.
On ImageNet, GRRE attains 91.6 / 98.6 (ACC/AUC) for both ADM and SD-v1 generators, outperforming DIRE and FIRE by a large margin (up to +25 \% ACC improvement over FIRE).
On LSUN-Bedroom, GRRE maintains near-perfect scores (98.6–100 / 99.9–100) across all diffusion models, while other methods such as AEROBLADE and FakeInversion exhibit notable degradation. 
These results indicate that GRRE generalizes effectively from facial domains to diverse indoor and object-centric scenes, confirming that the G-channel removed reconstruction captures universal artifact cues independent of content semantics.

In the reverse direction, where detectors are trained on LSUN-Bedroom (StyleGAN) and tested on CelebA-HQ and ImageNet, GRRE again shows superior robustness and domain transferability.
When transferred to CelebA-HQ, GRRE achieves perfect or near-perfect results (100 / 100 across all tested generators), whereas the performance of DIRE and FIRE drops significantly on challenging cases such as DALLE·2 and IF .
On ImageNet, GRRE still maintains high performance (85.2–91.6 / 99.7–100), notably outperforming FIRE and FakeInversion by 30–50 \% in accuracy.
Even when trained solely on GAN-generated images, GRRE remains capable of accurately detecting images synthesized by diffusion models, demonstrating strong cross-model transferability.
This strong cross-domain consistency demonstrates that GRRE effectively mitigates dataset bias and leverages channel-level reconstruction discrepancies that remain stable across heterogeneous domains.

Unlike traditional difference-based approaches that overfit to domain-specific textures, GRRE benefits from its color-channel perturbation design, which enhances sensitivity to intrinsic generative inconsistencies while preserving cross-domain invariance.
These results validate that GRRE not only performs excellently within individual datasets but also generalizes effectively across unseen domains and generator families, demonstrating outstanding robustness and adaptability in real-world scenarios.
\begin{figure}[t]
    \centering
fan fanfan fan t    \includegraphics[width=0.88\linewidth]{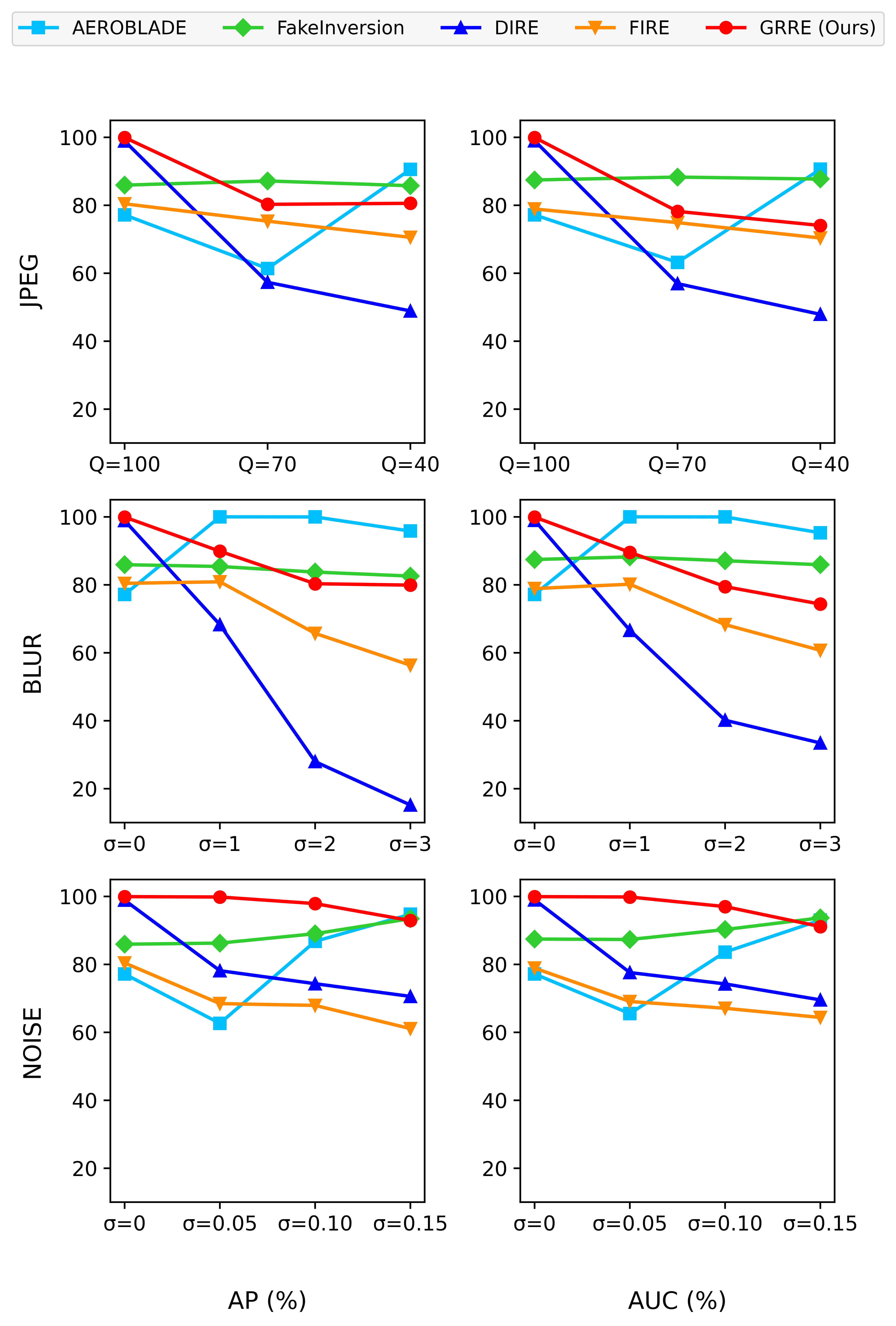}
    \caption{
        Detection robustness of GRRE and baselines under unseen perturbations,
        measured by AUC and AP across JPEG compression, Gaussian blur, and Gaussian noise.
    }
    \label{fig:robustness}
\end{figure}
\subsection{Robustness to Unseen Perturbations}
We further evaluate the robustness of the proposed GRRE framework against unseen image perturbations, including \textit{JPEG compression}, \textit{Gaussian blur}, and \textit{Gaussian noise}. As shown in \cref{fig:robustness}, GRRE consistently surpasses all baselines (\ie, DIRE, FIRE, FakeInversion, and AEROBLADE) in both \textbf{AUC} and \textbf{AP} metrics across all distortion types and intensity levels. Under JPEG compression, GRRE maintains high detection accuracy (AUC $\approx 74\%$) at quality factor $Q{=}40$ with minimal degradation, whereas competing methods exhibit substantial performance drops. When subjected to Gaussian blur, GRRE preserves stable discriminative capability even at standard deviation $\sigma{=}3$, while most baselines experience severe collapse. For Gaussian noise, GRRE achieves the highest overall scores across all perturbation levels, maintaining AUC above 90\% even at $\sigma{=}0.15$. These consistent advantages highlight the superior robustness of GRRE, which stems from its G-channel removed reconstruction error representation—effectively capturing texture-level inconsistencies that remain robust against pixel-level degradations—and its stable feature-domain representation that enables reliable generalization under various distortion conditions. Overall, GRRE demonstrates strong resistance to unseen perturbations, significantly outperforming prior detection methods in both accuracy and stability.
\subsection{More Analysis}
To further verify the rationality and superiority of the proposed \textbf{GRRE} framework, we compare it with two variant models: \textbf{RRRE}, which removes the red channel before reconstruction, and \textbf{BRRE}, which removes the blue channel. All three models share the same reconstruction–difference pipeline, differing only in the removed color channel. As summarized in \cref{tab:grre_vs_rrre_brre}, GRRE achieves the best performance across nearly all datasets and generative models. Specifically, GRRE attains an average AUC/AP of 98.44/99.79, significantly outperforming RRRE (81.84/95.08) and BRRE (87.62/99.56). The performance gain is particularly evident on diffusion-based models such as ADM and DDPM, where the removal of the G-channel yields more stable reconstruction residuals and richer structural discrepancies. These results confirm that the green channel—being the most information-dense component in the RGB space—serves as a more effective basis for reconstruction error modeling. Consequently, GRRE not only demonstrates superior discriminative ability but also validates the theoretical motivation behind channel-specific removal in our design. Notably, RRRE and BRRE also achieve competitive performance on certain models within the CelebA-HQ and ImageNet datasets, indicating that different color channels contribute complementary cues to generative artifact modeling.

We further conduct additional experiments on several recent, high-quality generative models (\textit{Imagen}\cite{Imagen}, \textit{Ideogram}\cite{Ideogram2023}, \textit{Kandinsky}\cite{kandinsky}, \textit{DALLE·3}\cite{dalle3}, \textit{PixArt}\cite{PixArt}, and \textit{SDXL}\cite{SDXL}) to evaluate the generalization ability of our method. 
The results are provided in the \textbf{Supplementary Material}.

\begin{table}[t]
    \caption{Comparison between GRRE and its channel-removal variants. 
    Each cell reports AUC/AP (\%) scores.
    }
    \label{tab:grre_vs_rrre_brre}
    \centering
    \resizebox{0.88\linewidth}{!}{
    \begin{tabular}{l l c c c}
        \toprule
        \textbf{Dataset} & \textbf{Model} & \textbf{GRRE (Ours)} & \textbf{RRRE} & \textbf{BRRE} \\
        \midrule
        \textbf{CelebA-HQ} & DALLE·2 & \cellcolor{blue!10}\textbf{100.0/100.0} & 100.0/100.0 & 99.95/100.0 \\
         & IF & 99.95/100.0 & 99.90/100.0 & \cellcolor{blue!10}\textbf{100.0/100.0} \\
         & Mid. & \cellcolor{blue!10}\textbf{100.0/100.0} & 100.0/100.0 & 99.95/100.0 \\
         & SD-v2 & \cellcolor{blue!10}\textbf{100.0/100.0} & 100.0/100.0 & 100.0/100.0 \\
        \midrule
        \textbf{ImageNet} & ADM & 91.60/98.61 & 99.50/99.98 & \cellcolor{blue!10}\textbf{99.67/100.0} \\
         & SD-v1 & 91.69/98.59 & \cellcolor{blue!10}\textbf{99.33/100.0} & 99.00/99.94 \\
        \midrule
        \textbf{LSUN-B.} & ADM & \cellcolor{blue!10}\textbf{98.65/99.93} & 68.00/93.34 & 82.50/99.46 \\
         & DDPM & \cellcolor{blue!10}\textbf{98.19/99.86} & 74.00/93.58 & 76.50/96.78 \\
         & IDDPM & \cellcolor{blue!10}\textbf{98.10/99.89} & 75.50/94.48 & 86.00/97.45 \\
         & PNDM & \cellcolor{blue!10}\textbf{95.75/99.61} & 76.50/95.94 & 92.50/99.90 \\
         & SD-v1 & \cellcolor{blue!10}\textbf{99.75/100.0} & 92.50/99.63 & 99.50/100.0 \\
         & SD-v2 & \cellcolor{blue!10}\textbf{100.0/100.0} & 77.50/97.37 & 93.50/100.0 \\
         & LDM & \cellcolor{blue!10}\textbf{99.95/100.0} & 57.00/84.31 & 84.00/99.94 \\
         & VQD & \cellcolor{blue!10}\textbf{99.95/100.0} & 53.50/75.25 & 64.00/99.96 \\
         & IF & \cellcolor{blue!10}\textbf{100.0/100.0} & 61.50/90.21 & 66.00/99.59 \\
         & DALLE·2 & \cellcolor{blue!10}\textbf{99.93/100.0} & 65.00/92.83 & 74.00/100.0 \\
         & Mid. & \cellcolor{blue!10}\textbf{100.0/100.0} & 91.00/99.45 & 72.50/99.55 \\
        \midrule
        \textbf{avg} &  & \cellcolor{blue!10}\textbf{98.44/99.79} & 81.84/95.08 & 87.62/99.56 \\
        \bottomrule
    \end{tabular}}
\end{table}
\begin{figure*}[t]
    \centering
    \includegraphics[width=0.88\textwidth]{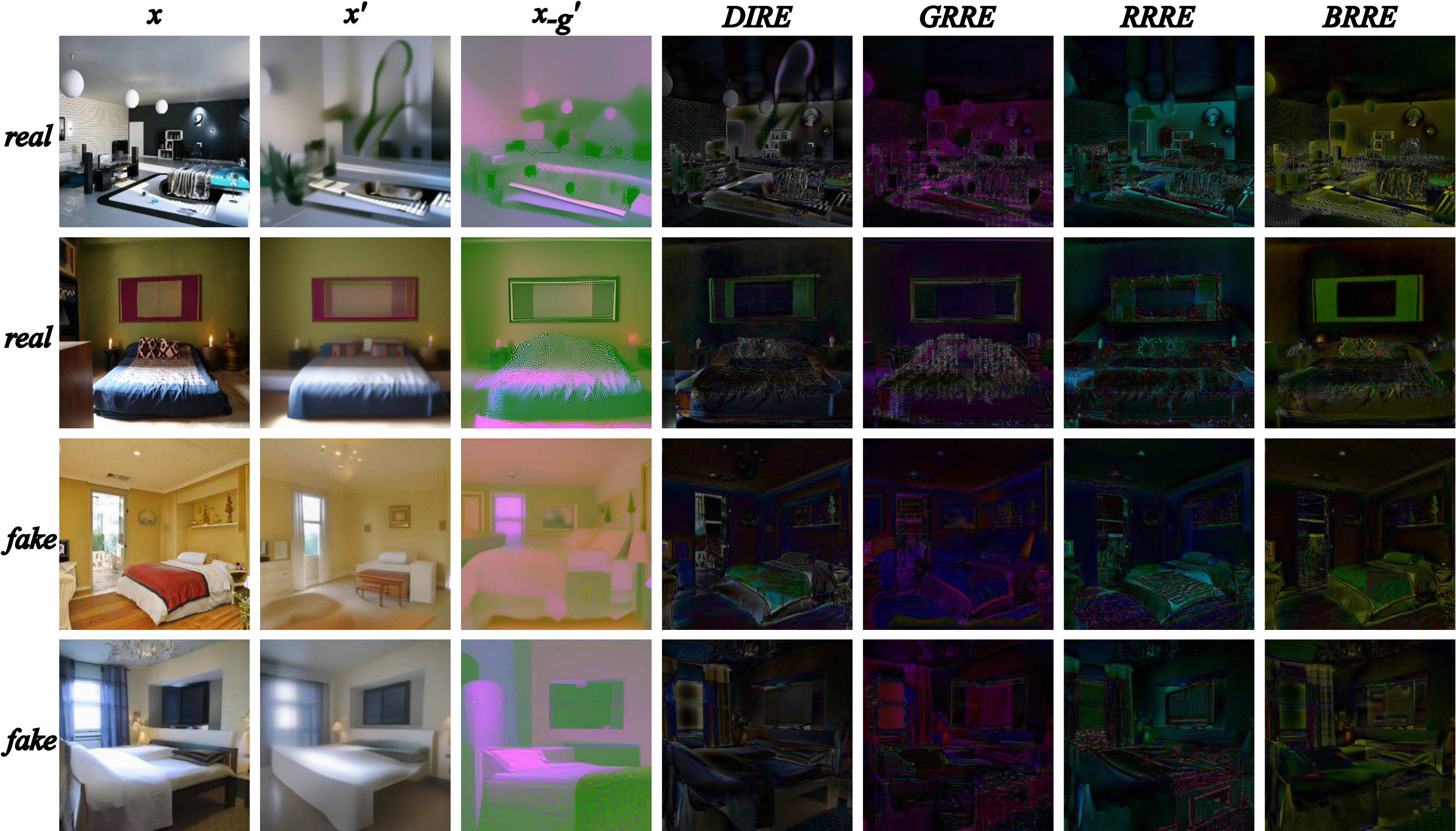}
    \caption{Visual comparison among DIRE, GRRE, RRRE, and BRRE on real and generated samples. 
    }
    \label{fig:all_visual}
\end{figure*}
\subsection{Visual Analysis}
To further demonstrate the interpretability and visual behavior of our proposed GRRE framework, we perform qualitative analyses on both real and generated samples. Two sets of visualization experiments are conducted to intuitively compare the reconstruction and residual differences between various methods.

Figure \ref{fig:all_visual} illustrates the visual comparison among the original image $x$, reconstructed image $x'$, G-channel removed reconstruction $x'_{-g}$, and the corresponding difference maps generated by DIRE and GRRE.
For real images, both DIRE and GRRE produce low-intensity residuals, indicating stable reconstruction.
However, for real images, the G-channel removed reconstruction becomes noticeably disordered, leading to stronger and more structured residual responses in the GRRE maps.
In contrast, DIRE exhibits weaker differentiation between real and fake samples.
This observation confirms that the G-channel removal effectively amplifies the underlying inconsistencies of generated images, thus enhancing the discriminative capability of GRRE.

To further explore the effect of different color-channel removal strategies, Figure \ref{fig:all_visual} compares the difference maps produced by GRRE (G-removed), BRRE (B-removed), RRRE (R-removed), and the baseline DIRE.
Across both real and fake images, all three channel-removed variants produce stronger residual responses than DIRE, verifying the benefit of channel perturbation.
Among them, GRRE consistently yields the most distinct and spatially coherent patterns, showing superior sensitivity to generative artifacts such as color inconsistencies and repetitive textures.
These visual results provide a clear and intuitive explanation for the quantitative performance advantage of GRRE.

To further analyze the visual behavior of our method, we include qualitative results in the \textbf{Supplementary Material}, where GRRE is applied to real and generated samples from the CelebA-HQ and ImageNet datasets.
These visualizations demonstrate that GRRE produces clearer and more structured residual patterns on real images while maintaining lower responses on generated ones, providing additional evidence of its robustness across domains.

\section{Conclusion}
In this work, we introduced GRRE (G-channel Removed Reconstruction Error), a novel reconstruction-based framework for detecting AI-generated images. By removing the green channel prior to reconstruction, GRRE effectively amplifies the inherent inconsistencies of real images while maintaining the structural integrity of synthetic ones. Extensive experiments across multiple datasets and generative models demonstrate that GRRE achieves state-of-the-art accuracy, robustness, and cross-domain generalization, outperforming existing reconstruction-based detectors. The proposed GRRE captures universal generative artifacts rather than dataset-specific cues, enabling consistent performance across different generators and domains. Looking ahead, we plan to extend GRRE by integrating multi-channel removal residuals—namely GRRE, RRRE, and BRRE—into a unified tri-channel fusion detector that leverages complementary color-space information. Furthermore, we will explore the feasibility of a training-free synthetic image detection paradigm, where multi-channel reconstruction errors can be adaptively fused without explicit supervision, paving the way toward more robust, interpretable, and deployable forensic systems against the rapidly evolving landscape of AI-generated imagery.
{
    \small
    \bibliographystyle{ieeenat_fullname}
    \bibliography{main}

@article{DDPMs,
  title={Denoising diffusion probabilistic models},
  author={Ho, Jonathan and Jain, Ajay and Abbeel, Pieter},
  journal={Advances in neural information processing systems},
  volume={33},
  pages={6840--6851},
  year={2020}
}

@inproceedings{iDDPM,
  title={Improved denoising diffusion probabilistic models},
  author={Nichol, Alexander Quinn and Dhariwal, Prafulla},
  booktitle={International conference on machine learning},
  pages={8162--8171},
  year={2021},
  organization={PMLR}
}

@inproceedings{LDMs,
  title={High-resolution image synthesis with latent diffusion models},
  author={Rombach, Robin and Blattmann, Andreas and Lorenz, Dominik and Esser, Patrick and Ommer, Bj{\"o}rn},
  booktitle={Proceedings of the IEEE/CVF conference on computer vision and pattern recognition},
  pages={10684--10695},
  year={2022}
}

@article{GANs,
  title={Generative adversarial nets},
  author={Goodfellow, Ian J and Pouget-Abadie, Jean and Mirza, Mehdi and Xu, Bing and Warde-Farley, David and Ozair, Sherjil and Courville, Aaron and Bengio, Yoshua},
  journal={Advances in neural information processing systems},
  volume={27},
  year={2014}
}

@article{DCGAN,
  title={Unsupervised representation learning with deep convolutional generative adversarial networks},
  author={Radford, Alec and Metz, Luke and Chintala, Soumith},
  journal={arXiv preprint arXiv:1511.06434},
  year={2015}
}

@article{ProGAN,
  title={Progressive growing of gans for improved quality, stability, and variation},
  author={Karras, Tero and Aila, Timo and Laine, Samuli and Lehtinen, Jaakko},
  journal={arXiv preprint arXiv:1710.10196},
  year={2017}
}

@inproceedings{StyleGAN,
  title={A style-based generator architecture for generative adversarial networks},
  author={Karras, Tero and Laine, Samuli and Aila, Timo},
  booktitle={Proceedings of the IEEE/CVF conference on computer vision and pattern recognition},
  pages={4401--4410},
  year={2019}
}

@inproceedings{StyleGAN2,
  title={Analyzing and improving the image quality of stylegan},
  author={Karras, Tero and Laine, Samuli and Aittala, Miika and Hellsten, Janne and Lehtinen, Jaakko and Aila, Timo},
  booktitle={Proceedings of the IEEE/CVF conference on computer vision and pattern recognition},
  pages={8110--8119},
  year={2020}
}

@article{SDEs,
  title={Score-based generative modeling through stochastic differential equations},
  author={Song, Yang and Sohl-Dickstein, Jascha and Kingma, Diederik P and Kumar, Abhishek and Ermon, Stefano and Poole, Ben},
  journal={arXiv preprint arXiv:2011.13456},
  year={2020}
}

@article{DMsGAN,
  title={Diffusion models beat gans on image synthesis},
  author={Dhariwal, Prafulla and Nichol, Alexander},
  journal={Advances in neural information processing systems},
  volume={34},
  pages={8780--8794},
  year={2021}
}

@inproceedings{DALL-E,
  title={Zero-shot text-to-image generation},
  author={Ramesh, Aditya and Pavlov, Mikhail and Goh, Gabriel and Gray, Scott and Voss, Chelsea and Radford, Alec and Chen, Mark and Sutskever, Ilya},
  booktitle={International conference on machine learning},
  pages={8821--8831},
  year={2021},
  organization={Pmlr}
}

@article{SDXL,
  title={Sdxl: Improving latent diffusion models for high-resolution image synthesis},
  author={Podell, Dustin and English, Zion and Lacey, Kyle and Blattmann, Andreas and Dockhorn, Tim and M{\"u}ller, Jonas and Penna, Joe and Rombach, Robin},
  journal={arXiv preprint arXiv:2307.01952},
  year={2023}
}

@article{Imagen,
  title={Imagen: Photorealistic text-to-image diffusion models},
  author={Saharia, Chitwan and Chan, William and Saxena, Saurabh and Li, Lala and Whang, Jay and Denton, Emily and Salimans, Tim and Ho, Jonathan and Fleet, David J and Norouzi, Mohammad},
  journal={arXiv preprint arXiv:2205.11487},
  volume={3},
  year={2022}
}

@inproceedings{PixArt,
  title={Pixart-$\sigma$: Weak-to-strong training of diffusion transformer for 4k text-to-image generation},
  author={Chen, Junsong and Ge, Chongjian and Xie, Enze and Wu, Yue and Yao, Lewei and Ren, Xiaozhe and Wang, Zhongdao and Luo, Ping and Lu, Huchuan and Li, Zhenguo},
  booktitle={European Conference on Computer Vision},
  pages={74--91},
  year={2024},
  organization={Springer}
}

@inproceedings{Sana,
  title={SANA: Efficient high-resolution text-to-image synthesis with linear diffusion transformers},
  author={Xie, Enze and Chen, Junsong and Chen, Junyu and Cai, Han and Tang, Haotian and Lin, Yujun and Zhang, Zhekai and Li, Muyang and Zhu, Ligeng and Lu, Yao and others},
  booktitle={The Thirteenth International Conference on Learning Representations},
  year={2025}
}

@inproceedings{SD3,
  title={Scaling rectified flow transformers for high-resolution image synthesis},
  author={Esser, Patrick and Kulal, Sumith and Blattmann, Andreas and Entezari, Rahim and M{\"u}ller, Jonas and Saini, Harry and Levi, Yam and Lorenz, Dominik and Sauer, Axel and Boesel, Frederic and others},
  booktitle={Forty-first international conference on machine learning},
  year={2024}
}

@misc{dalle2,
  author       = {Aditya Ramesh and others},
  title        = {DALL{\textperiodcentered}E 2: OpenAI Technical Preview},
  year         = {2022},
  note         = {OpenAI Technical Report},
}

@inproceedings{SD3-Turbo,
  title={Fast high-resolution image synthesis with latent adversarial diffusion distillation},
  author={Sauer, Axel and Boesel, Frederic and Dockhorn, Tim and Blattmann, Andreas and Esser, Patrick and Rombach, Robin},
  booktitle={SIGGRAPH Asia 2024 Conference Papers},
  pages={1--11},
  year={2024}
}

@article{FlowTok,
  title={Flowtok: Flowing seamlessly across text and image tokens},
  author={He, Ju and Yu, Qihang and Liu, Qihao and Chen, Liang-Chieh},
  journal={arXiv preprint arXiv:2503.10772},
  year={2025}
}

@inproceedings{DIRE,
  title={Dire for diffusion-generated image detection},
  author={Wang, Zhendong and Bao, Jianmin and Zhou, Wengang and Wang, Weilun and Hu, Hezhen and Chen, Hong and Li, Houqiang},
  booktitle={Proceedings of the IEEE/CVF International Conference on Computer Vision},
  pages={22445--22455},
  year={2023}
}

@inproceedings{LaRE2,
  title={LaRE\^{} 2: Latent reconstruction error based method for diffusion-generated image detection},
  author={Luo, Yunpeng and Du, Junlong and Yan, Ke and Ding, Shouhong},
  booktitle={Proceedings of the IEEE/CVF Conference on Computer Vision and Pattern Recognition},
  pages={17006--17015},
  year={2024}
}

@inproceedings{FIRE,
  title={Fire: Robust detection of diffusion-generated images via frequency-guided reconstruction error},
  author={Chu, Beilin and Xu, Xuan and Wang, Xin and Zhang, Yufei and You, Weike and Zhou, Linna},
  booktitle={Proceedings of the Computer Vision and Pattern Recognition Conference},
  pages={12830--12839},
  year={2025}
}

@article{PatchCraft,
  title={Patchcraft: Exploring texture patch for efficient ai-generated image detection},
  author={Zhong, Nan and Xu, Yiran and Li, Sheng and Qian, Zhenxing and Zhang, Xinpeng},
  journal={arXiv preprint arXiv:2311.12397},
  year={2023}
}

@article{CSFD,
  title={Optimized Frequency Collaborative Strategy Drives AI Image Detection},
  author={Li, Jun and Jiang, Wentao and Shen, Liyan and Ren, Yawei},
  journal={IEEE Internet of Things Journal},
  year={2025},
  publisher={IEEE}
}

@inproceedings{HFMF,
  title={HFMF: Hierarchical Fusion Meets Multi-Stream Models for Deepfake Detection},
  author={Mehta, Anant and McArthur, Bryant and Kolloju, Nagarjuna and Tu, Zhengzhong},
  booktitle={Proceedings of the Winter Conference on Applications of Computer Vision},
  pages={724--733},
  year={2025}
}

@inproceedings{FrequencyFeaturesGanDetection,
  title={Watch your up-convolution: Cnn based generative deep neural networks are failing to reproduce spectral distributions},
  author={Durall, Ricard and Keuper, Margret and Keuper, Janis},
  booktitle={Proceedings of the IEEE/CVF conference on computer vision and pattern recognition},
  pages={7890--7899},
  year={2020}
}

@inproceedings{GANFingerprints,
  title={Attributing fake images to gans: Learning and analyzing gan fingerprints},
  author={Yu, Ning and Davis, Larry S and Fritz, Mario},
  booktitle={Proceedings of the IEEE/CVF international conference on computer vision},
  pages={7556--7566},
  year={2019}
}

@article{TreeRingWatermarks,
  title={Tree-rings watermarks: Invisible fingerprints for diffusion images},
  author={Wen, Yuxin and Kirchenbauer, John and Geiping, Jonas and Goldstein, Tom},
  journal={Advances in Neural Information Processing Systems},
  volume={36},
  pages={58047--58063},
  year={2023}
}

@misc{StableSignature,
  author       = {Stability AI},
  title        = {Stable Signature: Watermarking for AI-Generated Images},
  year         = {2023},
  note         = {Technical Report, Stability AI Blog Release},
}

@inproceedings{FaceForensics++,
  title={Faceforensics++: Learning to detect manipulated facial images},
  author={Rossler, Andreas and Cozzolino, Davide and Verdoliva, Luisa and Riess, Christian and Thies, Justus and Nie{\ss}ner, Matthias},
  booktitle={Proceedings of the IEEE/CVF international conference on computer vision},
  pages={1--11},
  year={2019}
}

@article{LSUN,
  title={Lsun: Construction of a large-scale image dataset using deep learning with humans in the loop},
  author={Yu, Fisher and Seff, Ari and Zhang, Yinda and Song, Shuran and Funkhouser, Thomas and Xiao, Jianxiong},
  journal={arXiv preprint arXiv:1506.03365},
  year={2015}
}

@article{GenImage,
  title={Genimage: A million-scale benchmark for detecting ai-generated image},
  author={Zhu, Mingjian and Chen, Hanting and Yan, Qiangyu and Huang, Xudong and Lin, Guanyu and Li, Wei and Tu, Zhijun and Hu, Hailin and Hu, Jie and Wang, Yunhe},
  journal={Advances in Neural Information Processing Systems},
  volume={36},
  pages={77771--77782},
  year={2023}
}

@article{FakeBench,
  title={Fakebench: Probing explainable fake image detection via large multimodal models},
  author={Li, Yixuan and Liu, Xuelin and Wang, Xiaoyang and Lee, Bu Sung and Wang, Shiqi and Rocha, Anderson and Lin, Weisi},
  journal={IEEE Transactions on Information Forensics and Security},
  year={2025},
  publisher={IEEE}
}

@article{Diffusion-basedVisualArtCreation,
  title={Diffusion-based visual art creation: A survey and new perspectives},
  author={Wang, Bingyuan and Chen, Qifeng and Wang, Zeyu},
  journal={ACM Computing Surveys},
  volume={57},
  number={10},
  pages={1--37},
  year={2025},
  publisher={ACM New York, NY}
}

@inproceedings{GlazeProtectingArtists,
  title={Glaze: Protecting artists from style mimicry by $\{$Text-to-Image$\}$ models},
  author={Shan, Shawn and Cryan, Jenna and Wenger, Emily and Zheng, Haitao and Hanocka, Rana and Zhao, Ben Y},
  booktitle={32nd USENIX Security Symposium (USENIX Security 23)},
  pages={2187--2204},
  year={2023}
}

@inproceedings{JieHuaStyleFeatureExtracting,
  title={JieHua Paintings Style Feature Extracting Model using Stable Diffusion with ControlNet},
  author={Gu, Yujia and Li, Haofeng and Fang, Xinyu and Peng, Zihan and Peng, Yinan},
  booktitle={Proceeding of the 2024 5th International Conference on Computer Science and Management Technology},
  pages={349--353},
  year={2024}
}

@article{PDANetDiffusionPainting,
  title={A novel flexible identity-net with diffusion models for painting-style generation},
  author={Zhao, Yifei and Liang, Ziqi and Qiu, Yingrui and Wang, Xiaona},
  journal={Scientific Reports},
  volume={15},
  number={1},
  pages={27896},
  year={2025},
  publisher={Nature Publishing Group UK London}
}

@article{DeepfakeIdentityTheft,
  title={The New Identity Theft: Deepfakes and the Rise of Synthetic Impersonation Scams},
  author={Alexander, Alina},
  journal={Available at SSRN 5368947},
  year={2025}
}

@article{DeepfakeUnveiled,
  title={Deepfake technology unveiled: The commoditization of AI and its impact on digital trust},
  author={Popa, Claudiu and Pallath, Rex and Cunningham, Liam and Tahiri, Hewad and Kesavarajah, Abiram and Wu, Tao},
  journal={arXiv preprint arXiv:2506.07363},
  year={2025}
}

@techreport{DHSDeepfake2023,
  title        = {Increasing Threat of DeepFake Identities},
  institution  = {U.S. Department of Homeland Security},
  year         = {2023},
  note         = {Technical Report, U.S. Department of Homeland Security. Available online.},
}

@article{DeepfakeDetection&MultimediaForensics,
author = {Soni, Nishchal},
year = {2025},
month = {09},
pages = {36-39},
title = {Deepfake Detection and Multimedia Forensics: Investigating Synthetic Media, Image Forgery, and Video Manipulation in Cybercrime Cases},
volume = {9},
journal = {ARC Journal of Forensic Science},
doi = {10.20431/2456-0049.0902005}
}

@article{ColoeImageDenoisingGreenChannelPrior,
  title={Color Image Denoising Using The Green Channel Prior},
  author={Kong, Zhaoming and Yang, Xiaowei},
  journal={arXiv preprint arXiv:2402.08235},
  year={2024}
}

@article{ImageDenoisingGreenChannel,
  title={Image denoising using green channel prior},
  author={Kong, Zhaoming and Deng, Fangxi and Yang, Xiaowei},
  journal={IEEE Transactions on Image Processing},
  year={2025},
  publisher={IEEE}
}

@article{JointDenoisingDemosaickingGreenChannel,
  title={Joint denoising and demosaicking with green channel prior for real-world burst images},
  author={Guo, Shi and Liang, Zhetong and Zhang, Lei},
  journal={IEEE Transactions on Image Processing},
  volume={30},
  pages={6930--6942},
  year={2021},
  publisher={IEEE}
}

@book{FoundationVision,
  author    = {Brian A. Wandell},
  title     = {Foundations of Vision},
  year      = {1995},
  publisher = {Sinauer Associates},
  address   = {Sunderland, MA},
  note      = {ISBN 978-0-87893-853-7. Online version available from Stanford University.},
}

@inproceedings{aeroblade,
  title={Aeroblade: Training-free detection of latent diffusion images using autoencoder reconstruction error},
  author={Ricker, Jonas and Lukovnikov, Denis and Fischer, Asja},
  booktitle={Proceedings of the IEEE/CVF Conference on Computer Vision and Pattern Recognition},
  pages={9130--9140},
  year={2024}
}

@inproceedings{fakeinversion,
  title={Fakeinversion: Learning to detect images from unseen text-to-image models by inverting stable diffusion},
  author={Cazenavette, George and Sud, Avneesh and Leung, Thomas and Usman, Ben},
  booktitle={Proceedings of the IEEE/CVF Conference on Computer Vision and Pattern Recognition},
  pages={10759--10769},
  year={2024}
}

@inproceedings{vqdiffusion,
  title={Vector quantized diffusion model for text-to-image synthesis},
  author={Gu, Shuyang and Chen, Dong and Bao, Jianmin and Wen, Fang and Zhang, Bo and Chen, Dongdong and Yuan, Lu and Guo, Baining},
  booktitle={Proceedings of the IEEE/CVF conference on computer vision and pattern recognition},
  pages={10696--10706},
  year={2022}
}

@article{if,
  title={Photorealistic text-to-image diffusion models with deep language understanding},
  author={Saharia, Chitwan and Chan, William and Saxena, Saurabh and Li, Lala and Whang, Jay and Denton, Emily L and Ghasemipour, Kamyar and Gontijo Lopes, Raphael and Karagol Ayan, Burcu and Salimans, Tim and others},
  journal={Advances in neural information processing systems},
  volume={35},
  pages={36479--36494},
  year={2022}
}

@article{pndm,
  title={Pseudo numerical methods for diffusion models on manifolds},
  author={Liu, Luping and Ren, Yi and Lin, Zhijie and Zhao, Zhou},
  journal={arXiv preprint arXiv:2202.09778},
  year={2022}
}

@inproceedings{imagenet,
  title={Imagenet: A large-scale hierarchical image database},
  author={Deng, Jia and Dong, Wei and Socher, Richard and Li, Li-Jia and Li, Kai and Fei-Fei, Li},
  booktitle={2009 IEEE conference on computer vision and pattern recognition},
  pages={248--255},
  year={2009},
  organization={Ieee}
}

@article{kandinsky,
  title={Kandinsky: an improved text-to-image synthesis with image prior and latent diffusion},
  author={Razzhigaev, Anton and Shakhmatov, Arseniy and Maltseva, Anastasia and Arkhipkin, Vladimir and Pavlov, Igor and Ryabov, Ilya and Kuts, Angelina and Panchenko, Alexander and Kuznetsov, Andrey and Dimitrov, Denis},
  journal={arXiv preprint arXiv:2310.03502},
  year={2023}
}

@misc{dalle3,
  author       = {James Betker and Gabriel Goh and Li Jing and Tim Brooks and Jianfeng Wang and Linjie Li and Long Ouyang and Juntang Zhuang and Joyce Lee and Yufei Guo and others},
  title        = {Improving Image Generation with Better Captions},
  year         = {2023},
  note         = {OpenAI Technical Report, DALL{\textperiodcentered}E 3},
}

@misc{Ideogram2023,
  author       = {Ideogram, Inc.},
  title        = {Ideogram},
  year         = {2023},
  note         = {Model Release, Ideogram AI Website.},
}
}


\end{document}